\documentclass{intech}

\usepackage{devanagari}
\usepackage{xcolor}
\usepackage{colortbl}
\usepackage{multirow}
\usepackage{hyperref}
\hypersetup{
    colorlinks,%
    citecolor=blue,%
    filecolor=black,%
    linkcolor=red,%
    urlcolor=blue
}

\usepackage[font=small,labelfont=bf]{caption}
\addto\captionsenglish{}
\newcommand{\highlight}[3][black]{{\fboxsep3pt\colorbox{#2}{\color{#1} #3}}}

\usepackage{fancybox}
\booktitle{\textbf{Appeared in `Advances in Character Recognition'},  \textbf{Editor - Xiaoqing Ding}, \textbf{ISBN 978-953-51-0823-8} \hfill \textbf{Authors' copy}\hspace*{0.4cm} }%

\chaptertitle{Stroke-Based Cursive Character Recognition} 

\authors{K.C. Santosh\footnote{Corresponding author: Santosh.KC@inria.fr \hfill \textbf{Authors' copy, 2012}}}
\affiliation{INRIA Nancy Grand Est Research Centre}
\country{France}

\secondauthors{Eizaburo Iwata}
\secondaffiliation{Universal Robot Co. Ltd.}
\secondcountry{Japan}

\begin{document}

\maketitle

\section{Introduction}		
Human eye can see and read what is written or displayed either in natural handwriting or in printed format. The same
work in case the machine does is called handwriting recognition. Handwriting recognition can be broken down into two
categories:  off-line and on-line.

\begin{description}
\item [Off-line character recognition]-- Off-line character recognition takes a raster image from a scanner (scanned images of the paper documents), digital
camera or other digital input sources. The image is binarised based on for instance, color pattern (color or gray
scale) so that the image pixels are either 1 or 0.

\item [On-line character recognition]-- In on-line, the current information is presented to the system and recognition (of character or word) is carried out
at the same time. Basically, it accepts a string of $(x,y)$ coordinate pairs from an electronic pen touching a pressure
sensitive digital tablet.
\end{description}
In this chapter, we keep focusing on on-line writer independent cursive character recognition engine. In what follows, we explain the importance of on-line handwriting recognition over off-line, the necessity of writer independent system and the importance as well as scope of cursive scripts like Devanagari. Devanagari is considered as one of the known cursive scripts~\cite{palC04PR,jayadevanKPP11SMC}. However, we aim to include other scripts related to the current study.

\subsection{Why On-line?} 
With the advent of handwriting recognition technology since a few decades~\cite{plamondon00PAMI,arica01}, applications are challenging. 
For example, OCR is becoming an integral part of document scanners, and is used in many applications such as 
postal processing, script recognition, banking, security (signature verification, for instance) and language identification. 
In handwriting recognition, feature selection has been an important issue~\cite{duetrier96PR}. 
Both structural and statistical features as well as their combination have been widely used~\cite{heutteO98PRL,foggia99IVC}. 
These features tend to vary since characters' shapes vary widely. As a consequence, local structural properties like intersection of lines, number of holes, concave arcs, end points and junctions change time to time. These are mainly due to 
\begin{itemize}
\item \textit{deformations} can be from any range of shape variations including geometric transformation such as translation, rotation, scaling and even stretching; and
\item \textit{defects} yield imperfections due to printing, optics, scanning, binarisation as well as poor segmentation. 
\end{itemize}

In the state-of-the-art of handwritten character recognition, several different studies have shown that off-line handwriting recognition offers less classification rate compared to on-line~\cite{tappert90, plamondon00PAMI}. Furthermore, on-line data offers significant reduction in memory and therefore space complexity. Another advantage is that the digital pen or a digital form on a tablet device immediately transforms your handwriting into a digital representation that can be reused later without having any risk of degradation usually associated with ancient handwriting. Based on all these reasons, one can cite a few examples~\cite{boccignone93PR,doermannR95IJCV,viard05PRL,qiaoNY06PAMI} where they mainly focus on temporal information as well as writing order recovery from static handwriting image. On-line handwriting recognition systems provide interesting results.   

On-line character recognition involves the automatic conversion of stroke as it is written on a special digitizer or PDA, where a sensor picks up the pen-tip movements as well as pen-up/pen-down switching. Such data is known as digital ink and can be regarded as a dynamic representation of handwriting. The obtained signal is converted into letter codes which are usable within computer and character-processing applications.

\begin{figure}[h!]
\centering
\begin{tabular}{c}{\includegraphics[scale = 0.9]{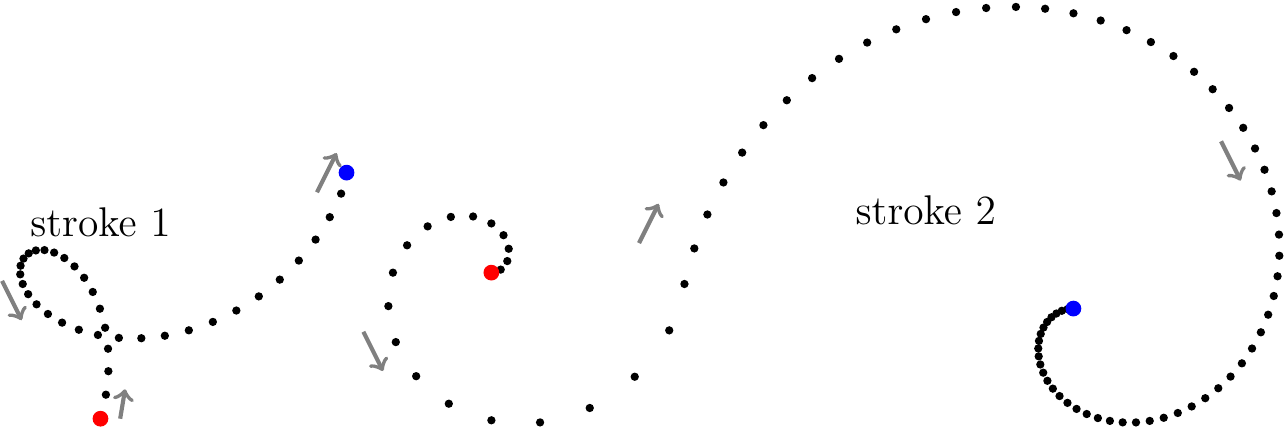}} \end{tabular}
\caption{On-line stroke sequences in the form of 2D $(x,y)$ coordinates. In this illustration, initial pen-tip position is coloured with red and pen-up (final point) is coloured with blue.}\label{stroke}
\end{figure}

The elements of an on-line handwriting recognition interface typically include:
\begin{enumerate}
\item a pen or stylus for the user to write with, and a touch sensitive surface, which may be integrated with, or adjacent to, an output display.
\item a software application i.e., a recogniser which interprets the movements of the stylus across the writing surface, translating the resulting strokes into digital character.
\end{enumerate}

Globally, it resembles one of the applications of pen computing i.e., computer user-interface using a pen (or stylus) and tablet, rather than devices such as a keyboard, joysticks or a mouse. Pen computing can be extended to the usage of mobile devices such as wireless tablet personal computers, PDAs and GPS receivers. 

Historically, pen computing (defined as a computer system employing a user-interface using a pointing device plus handwriting recognition as the primary means for interactive user input) predates the use of a mouse and graphical display by at least two decades, starting with the Stylator~\cite{dimond57} and RAND tablet~\cite{groner66} systems of the 1950s and early 1960s.


\subsection{Why Writer Independent?}
As mentioned before, on-line handwriting recognition systems provide interesting results almost over all types scripts. The recognition systems vary widely which can be due to nature of the scripts employed along with the associated particular difficulties including the intended applications. The performance of the application-based (commercial) recogniser is used to determine by its speed in addition to accuracy. 
 
Among many, more specifically, template based approaches have a long standing record~\cite{hu96hmm,schenkel95,connell99,bahlmann04,kc_pricai06}. In many of the cases, writer independent recogniser has been made since every new user does not require training -- which is widely acceptable. In such a context, the expected recognition system should automatically update or adapt the new users once they provide input or previously trained recogniser should be able to discriminate new users.   

\subsection{Why Devanagari?}\label{why devanagari}

In a few points, interesting scope will be summarised.
\begin{enumerate}
\item Pencil and paper can be preferable for anyone during a first draft preparation instead of using keyboard and other
computer input interfaces, especially when writing in languages and scripts for which keyboards are cumbersome.
Devanagari keyboards for instance, are quite difficult to use. Devanagari characters follow a complex structure and
may count up to more than 500 symbols~\cite{palC04PR,jayadevanKPP11SMC}.

\item Devanagari is a script used to write several Indian languages, including Nepali, Sanskrit, Hindi, Marathi, Pali,
Kashmiri, Sindhi, and sometimes Punjabi. According to the 2001 Indian census, 258 million people in India used
Devanagari.

\item Writing one's own style brings unevenness in writing units, which is the most difficult part to recognise. Variation
in basic writing units such as number of strokes, their order, shapes and sizes, tilting angles and similarities among
classes of characters are considered as the important issues. In contrast to Roman script, it happens more in cursive
scripts like Devanagari. 

Devanagari is written from left to right with a horizontal line on the \textit{top} which is the {\em shirorekha}. Every character requires one {\em shirorekha} from which text(s) is(are) suspended. The way of writing Devanagari has its own particularities. In what follows, in particular, we shortly explain a few major points associated difficulties.

\begin{itemize}
\item Many of the characters are similar to each other in structure. Visually very
similar {\em symbols} -- even from the same writer -- may represent different
{\em characters}. While it might seem quite obvious in the following examples to
distinguish the first from the second, it can easily be seen that confusion is likely to occur for their handwritten {\em symbol} counterparts ({\dn k}, {\dn P}), ({\dn y}, {\dn p}), ({\dn Y}, {\dn d}), etc.). Fig.~\ref{char. sample} shows a few examples of it.

\item The number of strokes, their order, shapes and sizes, directions, skew angle etc. are writing units that are important for symbol recognition
and classification. However, these writing units most often vary from one user to another and there is even no guarantee that a same user always writes in a same way. Proposed methods should take this into account.
\end{itemize}
\begin{figure}[tbp]
\centering
\includegraphics[scale=0.8]{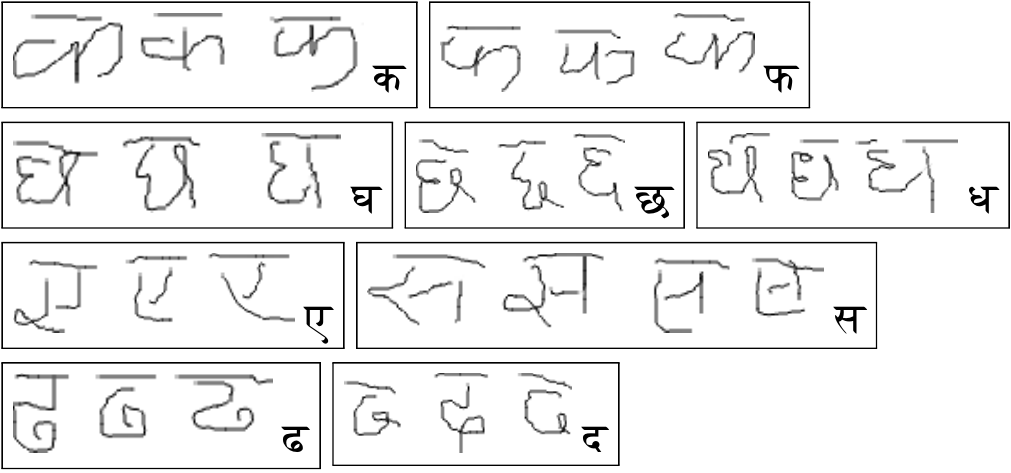}
\caption{A few samples of several different similar classes from Devanagari script.}\label{char. sample}
\end{figure}

\end{enumerate}
Based on those major aforementioned reasons, there exists clear motivation to pursue research on Devanagari handwritten character recognition.

\subsection{Structure of the Chapter}
The remaining of the paper is organised as follows. In Section~\ref{char. recog. framework}, we start with detailing the basic concept of character recognition framework in addition to the major highlights on important issues: feature selection, matching and recognition. Section~\ref{prop. recog. engine} gives a complete outline of how we can efficiently handle optimal recognition performance over cursive scripts like Devangari. In this section, we first provide the complete and then validate the whole process step by step with genuine reasoning and a series of experimental tests over our own dataset but, publicly available. We conclude the chapter in Section~\ref{conclusions}.

\section{Character Recognition Framework}\label{char. recog. framework}

Basically, we can categorise character recognition system into two modules: learning and testing. In learning or training module, following Fig.~\ref{learning}, handwritten strokes are learnt or stored. Testing module follows the former one. The performance of the recognition system is depends on how well handwritten strokes are learnt. It eventually refers to the techniques we employ. 

\begin{figure}[tbp]
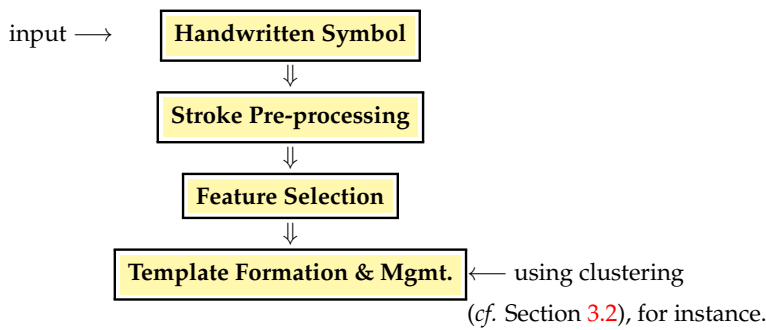

\centering
\renewcommand{\tabcolsep}{0em}
\begin{tabular}{rcl}
input $\longrightarrow$ & \framebox{\highlight{yellow!40}{\textbf{Handwritten Symbol}}}\\
&$\Downarrow$  \\
&\fbox{\highlight{yellow!40}{\textbf{Stroke Pre-processing}}} \\
&$\Downarrow$  \\
& \fbox{\highlight{yellow!40}{\textbf{Feature Selection}}} \\
& $\Downarrow$  \\
& \fbox{\highlight{yellow!40}{\textbf{Template Formation \& Mgmt.}}} & $\longleftarrow$ using clustering \\
& & (\textit{cf.} Section~\ref{clustering}), for instance.
\end{tabular}
\caption{Learning strokes from the handwritten symbols. In this illustration, we present a basic concept to form template via clustering of features of the strokes immediately after they are pre-processed.}\label{learning}
\end{figure}

\begin{figure}[tbp]
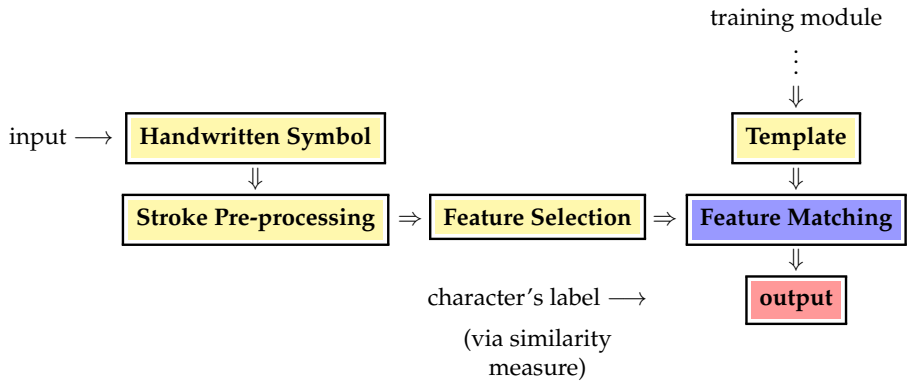

\centering
\begin{tabular}{rcccc}
&&& training module\\
&&& $\vdots$\\
&&& $\Downarrow$\\
input $\longrightarrow$& \framebox{\highlight{yellow!40}{\textbf{Handwritten Symbol}}} && \framebox{\highlight{yellow!40}{\textbf{Template}}} \\
&$\Downarrow$ && $\Downarrow$\\
&\fbox{\highlight{yellow!40}{\textbf{Stroke Pre-processing}}} & $\Rightarrow$ 
\fbox{\highlight{yellow!40}{\textbf{Feature Selection}}} $\Rightarrow$ &
\fbox{\highlight{blue!40}{\textbf{Feature Matching}}}\\
&&& $\Downarrow$\\
&& character's label $\longrightarrow$ & \fbox{\highlight{red!40}{\textbf{output}}}\\
&& (via similarity &\\
&& measure) &\\
\end{tabular}
\caption{An illustration of testing module. As in learning module, test characters are pre-processed and we present a basic concept to form template via clustering of features of the strokes immediately after they are pre-processed.}\label{testing}
\end{figure}

Basically, learning module employs stroke pre-processing, feature selection and clustering to form template to be stored. Pre-processing and feature selection techniques can be varied from one application to another. For example, noisy stroke elimination or deletion in Roman cannot be directly extended to the cursive scripts like Urdu and Devanagari. In other words, these techniques are found to be application dependent due to their different writing styles. However, they are basically adapted to each other and mostly ad-hoc techniques are built so that optimal recognition performance is possible. In the framework of stroke-based feature extraction and recognition, one can refer to~\cite{ChiuT99PR, Zhou:2007:ICDAR}, for example. It is important to notice that feature selection usually drives the way we match them. As an example, fixed size feature vectors can be straightforwardly matched while for non-linear feature vector sequences, dynamic programming (elastic matching) has been basically used~\cite{sakoe78ASSP,myers81,kruskall83,keogh99}. The concept was first introduced in the 60's~\cite{bellman59a}. Once we have an idea to find the similarity between the strokes' features, we follow clustering technique based on their similarity values. The clustering technique will generate templates as the representative of the similar strokes provided. These stored templates will be used for testing in the testing module. Fig.~\ref{testing} provides a comprehensive idea of it (testing module). More specifically, in this module, every test stroke will be matched with the templates (learnt in training module) so that we can find the most similar one. This procedure will be repeated for all available test strokes. At the end, aggregating all matching scores provides an idea of the test character closer to which one in the template.

\subsection{Preprocessing}
Strokes directly collected from users are often incomplete and noisy. Different systems use a variety of different pre-processing techniques before feature extraction~\cite{blumenstein03,verma04nn,yasser10CR}. The techniques used in one system may not exactly fit into the other because of different writing styles and nature of the scripts. Very common issues are repeated coordinates deletion~\cite{bahlmann04}, noise elimination and normalisation~\cite{guerfali93, chun05p}. 

Besides pre-processing, in this chapter, we mainly focus on feature selection and matching techniques. 

\subsection{Feature Selection}

If you have complete address of your friend then you can easily find him/her without an additional help from other people on the way. The similar case is happened in character recognition. Here, an address refers to a feature selection. Therefore, the complete or sufficient feature selection from the provided input is the crucial point. In other words, appropriate feature selection can greatly decrease the workload and simplify the subsequent design process of the classifier.

In what follows, we discuss a few but major issues associated with feature selection.
\begin{itemize}
\item Pen-flow i.e., speed while writing determines how well the coordinates along the pen trajectory are captured. Speed writing and writing with shivering hands, do not provide complete shape information of the strokes.

\item Ratios of the relative height, width and size 
of letters are not always consistent - which is obvious in natural handwriting. 

\item Pen-down and pen-up events provide stroke segmentation. But, we do not know which and where the strokes are rewritten or overwritten.

\item Slant writing style or writing with some angles to the left or right makes feature selection difficult. For example, in those cases, zoning information using orthogonal projection does not carry consistent information. This means that the zoning features will vary widely as soon as we have different writing styles.
\end{itemize}

We repeat, features should contain sufficient information to distinguish between classes, be insensitive to irrelevant variability of the input, allow efficient computation of discriminant functions and be able to limit the amount of training data required~\cite{lippmann89}. However, they vary from one script to another~\cite{blumenstein03,namboodiri04,verma04nn,okumura05}. 

\begin{figure}[tbp]
\centering
\includegraphics[scale=0.9]{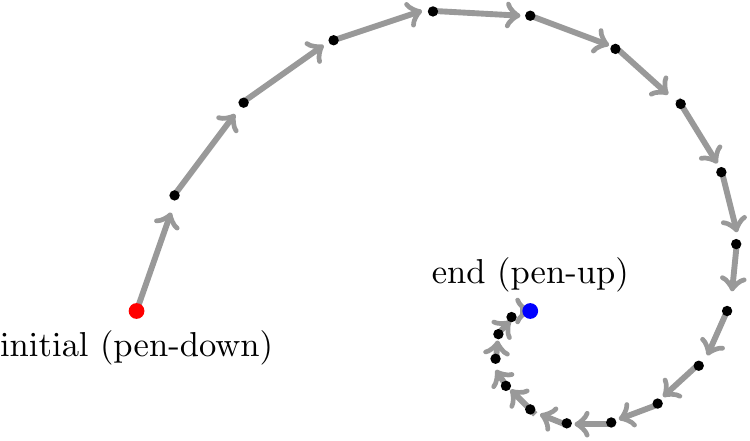}
\caption{An illustration of feature selection: pen-tip position and tangent at every pen-tip position along the pen trajectory.}\label{feature select.}
\end{figure}

Feature selection is always application dependent i.e., it relies on what type of scripts (their characteristics and difficulties) used. In our case, we use a feature vector sequence of any stroke is expressed as in~\cite{okumura05,kc_pricai06,kc_IJIG12}:
\begin{eqnarray} 
\mathbf{F} & = & \left[ \left( \mathbf{p}_{1},{\alpha}_{\mathbf{p}_{1},\mathbf{p}_{2}}\right), \left( \mathbf{p}_{2},{\alpha}_{\mathbf{p}_{2},\mathbf{p}_{3}}\right), \dots, \left( \mathbf{p}_{l-1},{\alpha}_{\mathbf{p}_{l-1},\mathbf{p}_{l}}\right) \right] \label{eq_features}
\end{eqnarray}
where, ${\alpha}_{\mathbf{p}_{l-1},\mathbf{p}_{l}} = \arctan\left(\frac{y_{l}-y_{l-1}} {x_{l}-x_{l-1}} \right)$. Fig.~\ref{feature select.} shows a complete illustration. 

Our feature includes a sequence of both pen-tip position and tangent angles sampled from the trajectory of the pen-tip, preserving the directional property of the trajectory path. It is important to remind that stroke direction (either left -- right or right -- left) leads to very different features although they are geometrically similar. To efficiently handle it, we need both kinds of strokes or samples for training and testing. This does not mean that same writer must be used.

The idea is somehow similar to the directional arrows that are composed of eight types, coded from $0-7$. This can be expressed
as, 
$
\begin{matrix}
\nwarrow   & \uparrow  & \nearrow\\
\leftarrow & \circ      & \rightarrow\\
\swarrow   & \downarrow & \searrow 
\end{matrix}
$.

However, these directional arrows provide only the directional feature of the strokes or line segments. Therefore, more information can be integrated if the relative length of the standard strokes is taken into account~\cite{ChaSS99ICDAR}.

\subsection{Feature Matching} \label{feature matching}
Besides, discussing on classifiers, we explain how features can be matched to obtain similarity or dissimilarity values between them.

Matching techniques are often induced by how features are taken or strokes are represented. For instance, normalising the feature vector sequence into a fixed size vector provides an immediate matching. On the other hand, features having different lengths or non-linear features need dynamic programming for approximate matching, for instance. Considering the latter situation, we explain how dynamic programming is employed.

Dynamic time warping (DTW) allows us to find the dissimilarity between two non-linear sequences potentially having different lengths~\cite{sakoe78ASSP,myers81,kruskall83,keogh99}. It is an algorithm particularly suited to matching sequences with missing information, provided there are long enough segments for matching to occur.

Let us consider two feature sequences 
\begin{eqnarray} \nonumber
\mathbf X &=& \{  \mathbf{x}_{k} \}_{k=1,\ldots,K} \mbox{ and} \\
\mathbf Y &=& \{ \mathbf{y}_{l} \}_{l=1,\ldots,L} \nonumber
\end{eqnarray} 
of size $K$ and $L$, respectively. The aim of the algorithm is to provide the optimal alignment between both sequences. At first, a matrix $M$ of size $K \times L$ is constructed. Then for each element in matrix $M$, local distance metric $\delta(k,l)$ between the events $e_k$ and $e_l$ is computed i.e., $\delta(k,l) = (e_k - e_l)^2$. Let $D(k,l)$ be the global distance up to $(k,l)$,
\begin{eqnarray}  \nonumber
D(k,l) &=&  \min \left[ \begin{matrix} D(k-1,l-1),\\ D(k-1,l), \\D(k,l-1)\end{matrix} \right]+ \delta(k,l)
\end{eqnarray}
with an initial condition $D(1,1) = \delta(1,1)$ such that it allows warping path going diagonally from starting node $(1,1)$ to end $(K,L)$. The main aim is to find the path for which the least cost is associated. The warping path therefore provides the difference cost between the compared signatures. Formally, the warping path is, 
\begin{eqnarray}\nonumber
{\cal W}  &=& \left\{ w_{t} \right\}_{t= 1 \ldots T},
\end{eqnarray} 
where $\mbox{ma}x(k,l) \leq T < k+l-1$ and $t^{th}$ element of $\cal{W}$ is $w(k,l)_t \in [1:K] \times [1: L]$ for $t \in [1:T]$. The optimised warping path $\cal{W}$ satisfies the following three conditions.
\begin{itemize}
\item [\textbf{c1.}] boundary condition: $$w_1 = (1,1) \mbox{ and } w_T = (K,L).$$
\item [\textbf{c2.}] monotonicity condition: $$k_1 \leq k_2 \leq \dots \leq k_K \mbox{ and } l_1 \leq l_2 \leq \dots \leq l_L.$$ 
\item [\textbf{c3.}] continuity condition: $$w_{t+1} - w_t \in \{ (1,1) (0,1), (1,0)\} \mbox{ for } t \in [1:T-1].$$   
\end{itemize}
\textbf{c1} conveys that the path starts from $(1,1)$ to $(K,L)$, aligning all elements to each other. \textbf{c2} forces the path advances one step at a time. \textbf{c3} restricts allowable steps in the warping path to adjacent cells, never be back. Note that \textbf{c3} implies \textbf{c2}.  

\begin{figure}[tbp]
\centering 
\begin{tabular}{c}
{\includegraphics[scale = 1.1]{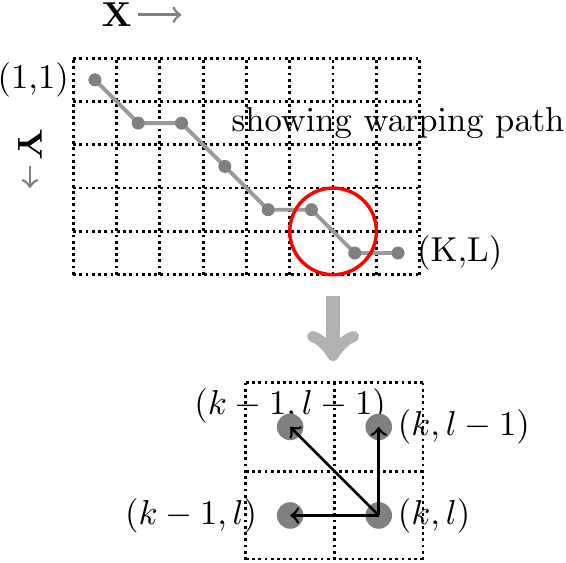}}\end{tabular}
\caption{Classical DTW algorithm -- an alignment illustration between two non-linear sequences $\mathbf{X}$ and $\mathbf{Y}$. In this illustration, diagonal DTW-matrix is shown including how back-tracking has been employed.}\label{dtw: algorithm}
\end{figure}

We then define the global distance between $\mathbf X$ and $\mathbf Y$ as,
\begin{eqnarray}\nonumber 
\Delta\left({\mathbf X},{\mathbf Y}\right) = \frac{D(K,L)}{T}.
\end{eqnarray}
The last element of the ${K \times L}$ matrix gives the DTW-distance between $\mathbf{X}$ and $\mathbf{Y}$, which is normalised by $T$ i.e., the number of discrete warping steps along the diagonal DTW-matrix. The overall process is illustrated in Fig.~\ref{dtw: algorithm}.

Until now, we provide a global concept of using DTW distance for non-linear sequences alignment. In order to provide faster matching, we have used local constraint on time warping proposed in~\cite{keogh02}. We have $w(k,l)_t$ such that $l-r \leq k \leq l+r$ where $r$ is a term defining a reach i.e., allowed range of warping for a given event in a sequence. With $r$, upper and lower bounding measures can be expressed as, 
\begin{eqnarray}\nonumber
\mbox{Upper bound } U_k = \mbox{max}(\mathbf x_{k-r} : \mathbf x_{k+r}) \\ \nonumber
\mbox{Lower bound } L_k = \mbox{min}(\mathbf x_{k-r} : \mathbf x_{k+r}). 
\end{eqnarray} 
Therefore, for all $i$, an obvious property of $U$ and $L$ is $U_k \geq \mathbf x_{k} \geq L_k$. With this, we can define a lower bounding measure for DTW: 
\begin{eqnarray}\nonumber
\mbox{LB\_Keogh}(\mathbf X, \mathbf Y) = \sqrt{\sum_{k=1}^{K} \left\{ \begin{array}{cc}
(\mathbf y_k - U_k)^2 & \mbox{if } \mathbf y_k > U_k \\
(\mathbf y_k - L_k)^2 & \mbox{if } \mathbf y_k < L_k\\
0 & \mbox{otherwise.}
\end{array}
\right. }
\end{eqnarray}
Since this provides a quick introduction of local constraint for lower bounding measure, we refer to~\cite{keogh02} for more clarification.

\subsection{Recognition}
From a purely combinatorial point of view, measuring the similarity or dissimilarity between two symbols 
$$\mathbf{S}_1= \left\{\mathbf{s}_1^i\right\}_{i=1\ldots n} \mbox{ and } \mathbf{S}_2 = \left\{\mathbf{s}_2^j\right\}_{j=1\ldots m}$$
composed, respectively, of $n$ and $m$ strokes, requires a one by one matching score computation of all strokes
$\mathbf{s}_1^i$ with all $\mathbf{s}_2^j$. This means that we align individual test strokes of an unknown symbols with the learnt strokes. As soon as we determine the test strokes associated with the known class, the complete symbol can be compared by the fusion of matching information from all test strokes. Such a concept is fundamental under the purview of stroke-based character recognition.

Overall, the concept may not always be sufficient, and these approaches generally need a final, global coherence check to avoid matching of strokes that shows visual similarity but do not respect overall geometric coherence within the complete handwritten character. In other words, matching strategy that happens between test stroke and templates of course, should be intelligent rather than straightforward one-to-many matching concepts. However, it in fact, depends on how template management has been made. In this chapter, this is one of the primary concerns. We highlight the use of relative positioning of the strokes within the handwritten symbol and its direct impact to the performance~\cite{kc_IJIG12}. 

\section{Recognition Engine}\label{prop. recog. engine}	
To make the chapter coherence as well as consistent (to Devanagari character recognition), it refers to the recognition engine which is entirely based on previous studies or works~\cite{kc_CIS06, kc_pricai06, kc_tujournal07, kc_icfhr10, kc_IJIG12}. Especially because of the structure of Devanagari, it is necessary to pay attention to the appropriate structuring of the strokes to ease and speed up comparison between the symbols, rather than just relying on global recognition techniques that would be based on a collection of strokes~\cite{kc_pricai06}. Therefore,~\cite{kc_icfhr10, kc_IJIG12} develop a method for analysing handwritten characters based on both the number of strokes and the their spatial information. It consists in four main phases.  

\begin{description}
\item [step 1.] Organise the symbols representing the same character into different groups based on the number of strokes.

\item [step 2.]  Find the spatial relation between strokes.

\item [step 3.]  Agglomerate similar strokes from a specific location in a group.

\item [step 4.]  Stroke-wise matching for recognition. 
\end{description}
For more clear understanding, we explain the aforementioned steps as follows. For a specific class of character, it is interesting to notice that writing symbols with the equal number of strokes, generally produce visually similar structure and is easier to compare. 

In every group within a particular class of character, a representative symbol is synthetically generated from pairwise similar strokes merging, which are positioned identically with respect to the {\em shirorekha}. It uses DTW algorithm. The learnt strokes are then stored accordingly. It is mainly focused on stroke clustering and management of the learnt strokes.

We align individual test strokes of an unknown symbols with the learnt strokes having both same number of strokes and spatial properties. Overall, symbols can be compared by the fusion of matching information from all test strokes. This eventually build a complete recognition process.

\subsection{Stroke Spatial Description and its Need}
The importance of the location of the strokes is best observed by taking a few pairs of characters that often lead to confusion:\\

({\huge{{\dn B}}} $\leftrightarrow$ {\huge{{\dn m}}}), ({\huge{{\dn D}}} $\leftrightarrow$ {\huge{{\dn G}}}), ({\huge{{\dn T}}} $\leftrightarrow$ {\huge{{\dn y}}}) etc. \\

The first character in every pair has visually two distinguishing features: its particular location of the \textit{shirorekha} (more to the right) and a small curve in the text. There is no doubt that one of the two features is sufficient to automatically distinguish both characters. However, small curves are usually not robust feature in natural handwriting, finding the location of the {\em shirorekha} only can avoid possible confusion. Our stroke based spatial relation technique is explained further in the following.

To handle relative positioning of strokes, we use six spatial predicates i.e., $2\times3$ relational regions: 
\begin{eqnarray} \nonumber
{\cal{R}} &= &
\left[
\begin{array}{lll}
\textit{top-left} \mbox{ (T--L)} & \hspace*{0.5em} \textit{top} \mbox{ (T)} & \hspace*{0.5em} \textit{top-right} \mbox{ (T--R)}\\
\textit{bottom-left} \mbox{ (B--L)} & \hspace*{0.5em} \textit{bottom} \mbox{ (B)} & \hspace*{0.5em} \textit{bottom-right} \mbox{ (B--L)}
\end{array}
\right].
\end{eqnarray}

For easier understanding, iconic representation of the aforementioned relational matrix $\cal R$ can be expressed as, 
$$
\begin{bmatrix} \circ & \circ & \circ \\ \circ & \circ & \bullet  \end{bmatrix}
$$ where black-dot represents the presence i.e., stroke is found to be in the provided \textit{bottom-right} region.

To confirm the location of the stroke, we use the projection theory: minimum boundary rectangle (MBR)~\cite{papadias94t} model combined with the stroke's centroid. 

Based on~\cite{egenhofer91}, we start with checking fundamental topological relations such as \textit{disconnected} (DC), \textit{externally connected} (EC) and \textit{overlap/intersect} (O/I) by considering two strokes $\mathbf{s^{j}}$ and $\mathbf{s^{j'}}$:
$$\mathbf{s^{j}} = \left\{\mathbf{p}_{k}^{j}\right\}_{k=1\ldots l} \mbox{ and } \mathbf{s^{j'}} = \left\{\mathbf{p}_{k'}^{j'}\right\}_{k'=1\ldots l'}$$ as follows, 
\begin{eqnarray}\nonumber
\mathbf{s^{j}}\cap \mathbf{s^{j'}} &=& \left\{ \begin{array}{ll}
1 & \mbox{if $(\mathbf{p}_{k}^{j} \cap \mathbf{p}_{k'}^{j'} \neq \emptyset)$ $\Rightarrow$ EC, O/I}\\
0 & \mbox{otherwise $\Rightarrow$ DC}.
\end{array}
\right.
\end{eqnarray}
We then use the border condition from the geometry of the MBR. It is straightforward for \textit{disconnected} strokes while, is not for \textit{externally connected} and \textit{overlap/intersect} configurations. In the latter case, we check the level of the centroid with respect to the boundary of the MBR. For example, if a boundary of the \textit{shirorekha} is above the centroid level of the \textit{text} stroke, then it is confirmed that the \textit{shirorekha} is on the \textit{top}. This procedure is applied to all of the six previously mentioned spatial predicates. Note that use of angle-based model like bi-centre~\cite{miyajima94a} and angle histogram~\cite{wang99a} are not the appropriate choice due to the cursive nature of writing.
\begin{figure}[tbp]
\centering
\begin{tabular}{cc}
\begin{tabular}{c}{\includegraphics[scale = 0.55]{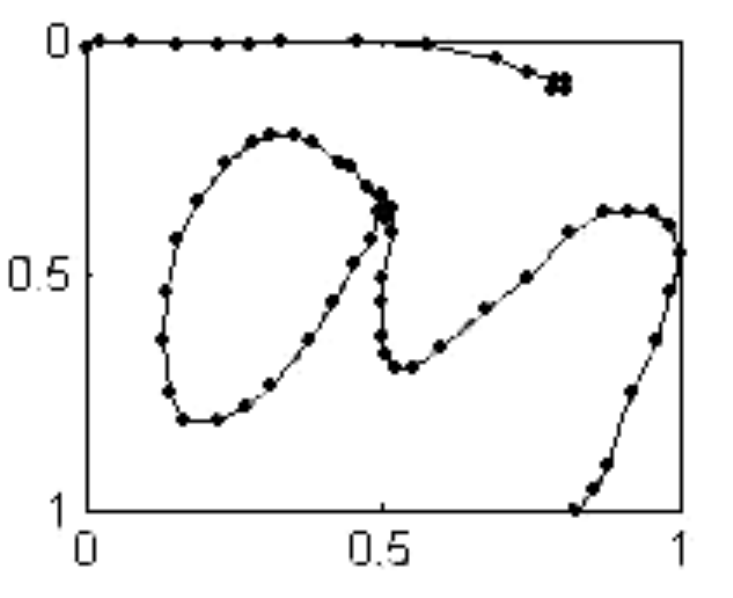}}\\ (a) Two-stroke {\textbf{\dn k}} \end{tabular} & $\Longrightarrow$
\begin{tabular}{c}{\includegraphics[scale = 0.6]{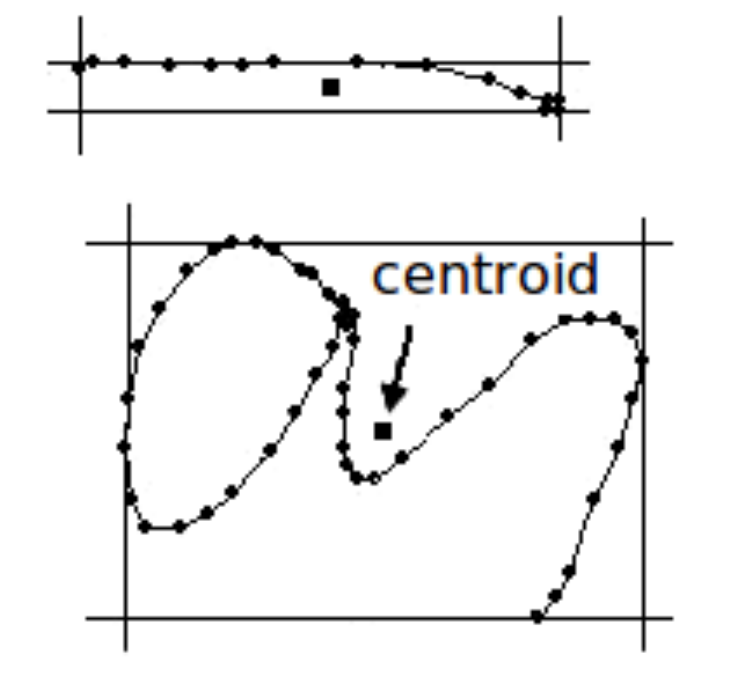}} \\ (b) MBR + Centroid \\ model \end{tabular}\\
&  \\[-0.5em] 
&$\Downarrow$\\[0.5em]
& \hspace*{0.6cm}
\begin{tabular}{c}{\includegraphics[scale = 0.55]{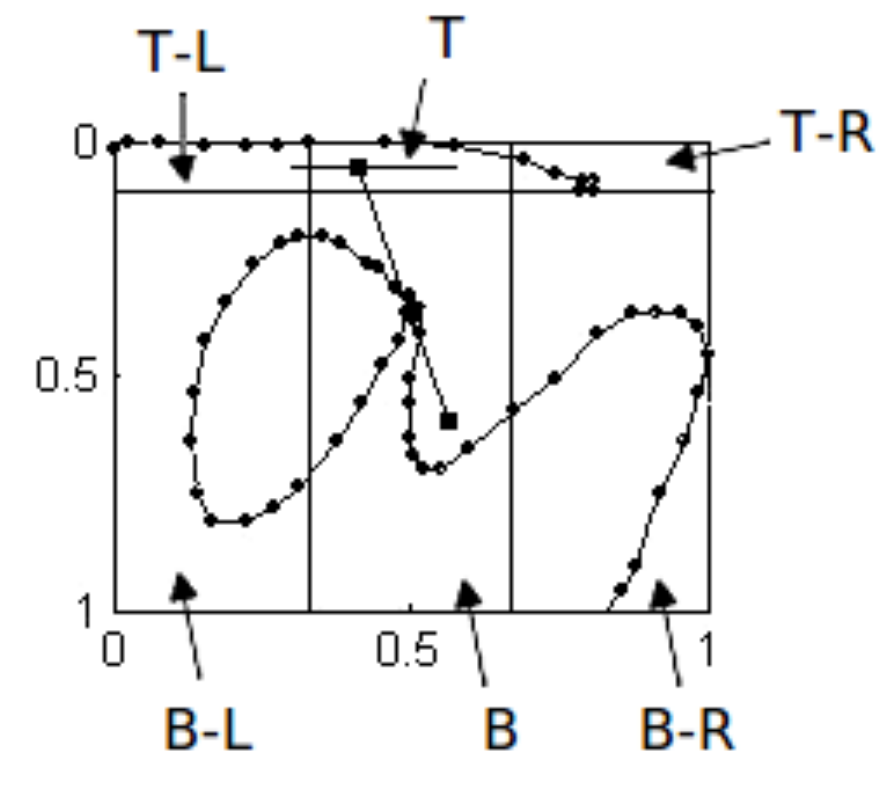}}\\ (c) Model realisation \end{tabular}
\end{tabular}
\caption{\label{spatial-relation} Pairwise spatial relation for a two-stroke {\dn k}.}
\end{figure}

On the whole, assuming that the {\em shirorekha} is on the \textit{top}, the locations of the \textit{text} strokes are estimated. This eventually allows to cross-validate the location of the {\em shirorekha} along with its size, once \textit{texts}' locations are determined. Fig.~\ref{spatial-relation} shows a real example demonstrating relative positioning between the strokes for a two-stroke symbol {\dn k}. Besides, symbols with two {\em shirorekha}s are also possible to treat. In such a situation, the first {\em shirorekha} according to the order of strokes is taken as reference.

\subsection{Spatial Similarity based Clustering}\label{clustering}
 Basically, clustering is a technique for collecting items which are similar in some way. Items of one group are dissimilar with other items belonging to other groups. Consequently, it makes the recognition system compact. To handle this, we present spatial similarity based stroke clustering.
\begin{figure}[tbp]
\centering
\begin{tabular}{c}
\begin{tabular}{c}{\includegraphics[width=0.53\linewidth]{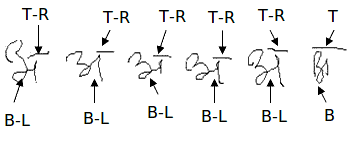}} \\ (a) Two-stroke {\dn a} \end{tabular}\\[2pt]
\begin{tabular}{c}{\includegraphics[width=0.53\linewidth]{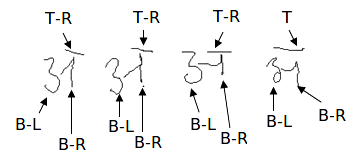}} \\ (b) Three-stroke {\dn a} \end{tabular} 
\end{tabular}
\caption{Relative positions of strokes for a class {\dn a} in two different groups i.e., two-stroke and three-stroke symbols.}\label{group}
 \end{figure}

\begin{figure}[tbp]
\centering
\begin{tabular}{c}
\begin{tabular}{c}{\includegraphics[scale = 1]{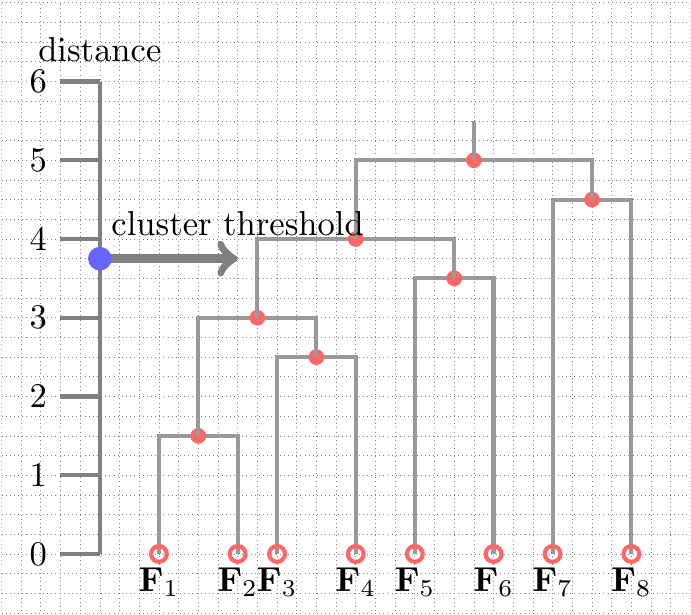}} \end{tabular}
\end{tabular}
\caption{Hierarchical stroke clustering concept. At every step, features are merged according to their similarity up to the provided threshold level.}\label{stroke clustering}
\end{figure}

As mentioned in previous work~\cite{kc_icfhr10, kc_IJIG12}, the clustering scheme  is a two-step process. 
\begin{itemize}
\item The first step is to organise symbols representing a same character into different groups, based on the number of strokes used to complete the symbol. Fig.~\ref{group} shows an example of it for a class of character {\dn a}. 

\item In the second step, strokes from the specific location are agglomerated hierarchically within the particular group. Once relative position for every stroke is determined as shown in Fig.~\ref{group}, single-linkage agglomerative hierarchical clustering is used (\textit{cf.} Fig.~\ref{stroke clustering}). This means that only strokes which are at a specific location are taken for clustering. As an example, we illustrate it in Fig.~\ref{clustering1}. This applies to all groups within a class. 
\end{itemize}

\begin{figure}[tbp]
\centering
\begin{eqnarray}\nonumber
\underbrace{\underbrace{\begin{bmatrix} \circ & \circ & \circ \\ \bullet & \circ & \circ  \end{bmatrix}}_{1}, \underbrace{\begin{bmatrix} \circ & \circ & \circ \\ \circ & \bullet & \circ  \end{bmatrix}}_{2}, \underbrace{\begin{bmatrix} \circ & \circ & \circ \\ \circ & \bullet & \circ  \end{bmatrix}}_{3}}_{text\mbox{ clustering}}  \mbox{ and } \underbrace{\underbrace{\begin{bmatrix} \bullet & \circ & \circ \\ \circ & \circ & \circ  \end{bmatrix}}_{1}, \underbrace{\begin{bmatrix} \circ & \bullet & \circ \\ \circ & \circ & \circ  \end{bmatrix}}_{2}, \underbrace{\begin{bmatrix} \circ & \circ & \bullet \\ \circ & \circ & \circ  \end{bmatrix}}_{3}}_{shirorekha\mbox{ clustering}}
\end{eqnarray}
\caption{Clustering technique for each class. Stroke clustering is based on the relative positioning. As a consequence, we have three clustering blocks for \textit{text} strokes and remaining three for \textit{shirorekha}.}\label{clustering1}
\end{figure}

In agglomerative hierarchical clustering (\textit{cf.} Fig.~\ref{stroke clustering}), we merge two similar strokes and find a new cluster. The distance computation between two strokes follows Section~\ref{feature matching}. The new cluster is computed by averaging both strokes via the use of the discrete warping path along the diagonal DTW-matrix. This process is repeated until it reaches the cluster threshold. The threshold value yields the number of cluster representatives i.e., learnt templates.

\subsection{Stroke Number and Order Free Recognition}\label{stroke nof}
\begin{figure}[tbp]
\centering
\begin{tabular}{c}{\includegraphics[height = 1.5cm, width = 1.8cm]{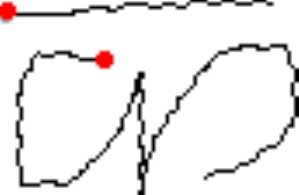}} \\ (a) two-stroke {\dn k}\end{tabular}\hspace*{4pt}
\begin{tabular}{c}{\includegraphics[height = 1.5cm, width = 1.8cm]{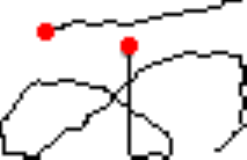}} \\ (b) two-stroke {\dn k}\end{tabular}\hspace*{4pt}
\begin{tabular}{c}{\includegraphics[height = 1.5cm, width = 1.8cm]{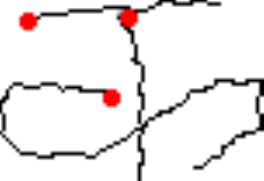}} \\ (c) three-stroke {\dn k}\end{tabular}\\[2pt]
\begin{tabular}{c}{\includegraphics[height = 1.5cm, width = 1.8cm]{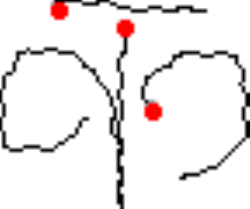}} \\ (d) three-stroke {\dn k}\end{tabular}\hspace*{4pt}
\begin{tabular}{c}{\includegraphics[height = 1.5cm, width = 1.8cm]{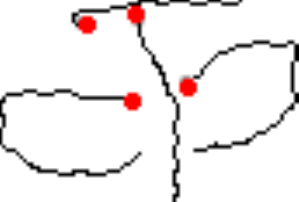}} \\ (e) four-stroke {\dn k}\end{tabular}\hspace*{4pt}
\begin{tabular}{c}{\includegraphics[height = 1.5cm, width = 1.8cm]{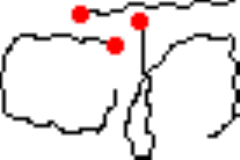}} \\ (f) three-stroke {\dn k}\end{tabular}
\caption{Different number of strokes and order for a class \textbf{{\dn k}}. In this illustration, red-dot refers to the initial pen-tip position so that it makes easy to realise how many number of strokes to make a complete symbol. In addition, stroke ordering is different from one to another.}\label{sample nof}
\end{figure}

In natural handwriting, number of strokes as well as their order vary widely. This happens from one writing to another, even from the same user -- which of course exits from different users. Fig.~\ref{sample nof} shows the large variation of stroke numbers as well as the orders.

Once we have organised the symbols (from the particular class) into groups based on the number of strokes used, our stroke clustering has been made according to the relative positioning. As a consequence, while doing recognition, one can write symbol with any numbers and orders because stroke matching is based on relative positioning of the strokes in which group while it does not need to care about the strokes order.  

\subsection{Dataset}
In this work, as before, publicly available dataset has been employed (\textit{cf.} Table~\ref{table_data}) where a Graphite tablet (WCACOM Co. Ltd.), model ET0405A-U, was used to capture the pen-tip position in the form of $2D$ coordinates at the
sampling rate of 20 Hz. The data set is composed of 1800 symbols representing
36 characters, coming from 25 native speakers. Each writer was given
the opportunity to write each character twice. No other directions,
constraints, or instructions were given to the users. 

\begin{table}[tbp]
\renewcommand{\tabcolsep}{1em}
\centering
\caption{Dataset formation and its availability.}\label{table_data}
\begin{tabular}{|ll|}
\hline
\rowcolor[gray]{.60}Item  & Description \\ \hline \hline
\rowcolor[gray]{.95}Classes of character & 36\\
\rowcolor[gray]{.85}Users & 25 \\
\rowcolor[gray]{.95}Dataset size & 1800 \\
\rowcolor[gray]{.85}Visibility & IAPR tc--11 \\
\rowcolor[gray]{.85} & \url{http://www.iapr-tc11.org}\\ \hline
\end{tabular}
\end{table}

\subsection{Recognition Performance Evaluation}
While experimenting, every test sample is matched with training candidates and the closest one is reported. The closest candidate corresponds to the labelled class, which we call `character recognition'. Formally, recognition rate can be defined as the number of correctly recognised candidates to the total number of test candidates.

To evaluate the recognition performance, two different protocols can be employed:
\begin{enumerate}
\item dichotomous classification and
\item ${\mathbb{K}}$-fold cross-validation (CV).
\end{enumerate}
In case of dichotomous classification, 15 writers are used for training and the remaining 10 are for testing. On the other hand, ${\mathbb{K}}$-fold CV has been implemented. Since we have 25 users for data collection, we employ $\mathbb{K} = 5$ in order to make recognition engine writer independent.

In ${\mathbb{K}}$-fold CV, the original sample for every class is randomly partitioned into ${\mathbb{K}}$ sub-samples. Of the $\mathbb{K}$ sub-samples, a single sub-sample is used for validation, and the remaining ${\mathbb{K}}-1$ sub-samples are used for training. This process is then repeated for ${\mathbb{K}}$ folds, with each of the ${\mathbb{K}}$ sub-samples used exactly once. Finally, a single value results from averaging all. The aim of the use of such a series of rigorous tests is to avoid the biasing of the samples that can be possible in conventional dichotomous classification. In contrast to the previous studies~\cite{kc_IJIG12}, this will be an interesting evaluation protocol.

\subsection{Results and Discussions}
Following evaluation protocols we have mentioned before, Table~\ref{results0} provides average recognition error rates. In the tests, we have found that the recognition performance has been advanced by approximately more than 2\%. 
 
Based on results (\textit{cf.} Table~\ref{results0}), we investigate the recognition performance based on the observed errors. We categorise the origin of the  errors that are occurred in our experiments. As said in Section~\ref{why devanagari}, these are mainly due to
\begin{enumerate}
\item structure similarity,
\item reduced and/or very long ascender and/or descender stroke, and
\item others such as re-writing strokes and mis-writing.
\end{enumerate}

Compared to previous work~\cite{kc_IJIG12}, number of rejection does not change while confusions due to structure similarity has been reduced. This is mainly because of the 5-fold CV evaluation protocol. Besides, running time has been reduced by more than a factor of two i.e., 2 seconds per character, thanks to LB\_Keogh tool~\cite{keogh02}.
\renewcommand{\tabcolsep}{0.4em}
\begin{table}[tbp]
\caption{Error rates (in \%) and running time (in sec. per character). The methods can be differentiated by the additional use of L\_B Keogh tool~\cite{keogh02} and the evaluation protocol employed.}\label{results0}
\centering
\begin{tabular} {|lllll|}
\hline
\rowcolor[gray]{.60}& {{$\#$ of}} & {$\#$ of} & {Avg.}& {Time} \\[-0.5pt]
\rowcolor[gray]{.60}{{Method}}  &  {Mis-recognition} & {{Rejection}} & {Error \%} & {sec.}\\[0.1em]
\hline \hline
\rowcolor[gray]{.95}M1.  & 33  & 08  & 05.0  & 04\\
\rowcolor[gray]{.85}M2.  & 24  & 08  & 03.5  & 02\\ \hline
\end{tabular}
\begin{tabular}{l}
Index:\\
M1.~\cite{kc_IJIG12}. \\
M2.~\cite{kc_IJIG12} +~\cite{keogh02} and 5-fold CV.
\end{tabular}
\end{table}

\section{Conclusions}\label{conclusions}		
In this chapter, an established as well as validated approach (based on previous studies~\cite{kc_CIS06, kc_pricai06, kc_tujournal07, kc_icfhr10, kc_IJIG12}) has been presented for on-line natural handwritten Devanagari character recognition. It uses the number of strokes used to complete a symbol and their spatial relations\footnote{A comprehensive work based on relative positioning of the handwritten strokes, is presented in~\cite{kc_IJIG12}. Once again, to avoid contradictions, this chapter aims to provide coherence as well as consistent studies on Devanagari character recognition.}. Besides, we have provided the dataset publicly available for research purpose. Considering such a dataset, the success rate is approximately $97\%$ in less than 2 seconds per character on average. In this chapter, note that the new evaluation protocol reduces the errors (mainly due to multi-class similarity) and the optimised DTW reduces the delay in processing -- which has been new attestation in comparison to the previous studies. 

The proposed approach is able to handle handwritten symbols of any stroke and order. Moreover, the stroke-matching technique is interesting and completely controllable. It is primarily due to our symbol categorisation and the use of stroke spatial information in template management. To handle spatial relation efficiently (rather than not just based on orthogonal projection i.e., MBR), more elaborative spatial relation model can be used~\cite{kc_11PRL}, for instance. In addition, use of machine learning techniques like inductive logic programming (ILP)~\cite{kc_LR09ICDAR, Amin00IJIS} to exploit the complete structural properties in terms of first order logic (FOL) description.

\section*{Acknowledgements}
Since the chapter is based on the previous studies, thanks to researchers Cholwich Nattee, School of ICT, SIIT, Thammasat University, Thailand and Bart Lamiroy, Universit\'e de Lorraine -- Loria Campus Scientifique, France for their efforts. Besides, the dataset is partially based on master thesis: TC-MS-2006-01, conducted in Knowledge Information \& Data Management Laboratory, School of ICT, SIIT, Thammasat University under Asian Development Bank -- Japan Scholarship Program (ADB-JSP).

\bibliography{biblio}

\end{document}